\documentclass[twocolumn]{article}
\usepackage{PRIMEarxiv}

\usepackage[utf8]{inputenc} 
\usepackage[T1]{fontenc}    
\usepackage{hyperref}       
\usepackage{url}            
\usepackage{booktabs}       
\usepackage{bm}
\usepackage{amsfonts}       
\usepackage{amsthm}
\usepackage{amsmath}
\usepackage{amssymb}
\usepackage{mathtools}
\usepackage{microtype}      
\usepackage{lipsum}
\usepackage{fancyhdr}       
\usepackage{graphicx}       
\graphicspath{{media/}}     
\usepackage{comment}
\usepackage{here}
\usepackage{color}
\usepackage{multirow}
\usepackage{hhline}
\usepackage{makecell}

\usepackage{tabularx}

\usepackage[capitalize]{cleveref}
\crefname{section}{Sec.}{Secs.}
\Crefname{section}{Section}{Sections}
\Crefname{table}{Table}{Tables}
\crefname{table}{Tab.}{Tabs.}

\title{Vision and Language Reference Prompt into SAM \\ for Few-shot Segmentation}

\author{
  Kosuke Sakurai \\
  Waseda University \\
  \texttt{kosukesakurai@toki.waseda.jp} \\\
\And
  Ryotaro Shimizu \\
  Waseda University \\
  \texttt{shi3mizu8-r@fuji.waseda.jp} \\\
\And
  Masayuki Goto \\
  Waseda University \\
  \texttt{masagoto@waseda.jp} \\
}

\begin{document}
\twocolumn[
\maketitle

\vspace{-5mm}
\begin{abstract}
\vspace{3mm}
   Segment Anything Model (SAM) represents a large-scale segmentation model that enables powerful zero-shot capabilities with flexible prompts. While SAM can segment any object in zero-shot, it requires user-provided prompts for each target image and does not attach any label information to masks. Few-shot segmentation models addressed these issues by inputting annotated reference images as prompts to SAM and can segment specific objects in target images without user-provided prompts. Previous SAM-based few-shot segmentation models only use annotated reference images as prompts, resulting in limited accuracy due to a lack of reference information. In this paper, we propose a novel few-shot segmentation model, \textit{Vision and Language reference Prompt into SAM} (VLP-SAM), that utilizes the visual information of the reference images and the semantic information of the text labels by inputting not only images but also language as reference information. In particular, VLP-SAM is a simple and scalable structure with minimal learnable parameters, which inputs prompt embeddings with vision-language information into SAM using a multimodal vision-language model. To demonstrate the effectiveness of VLP-SAM, we conducted experiments on the PASCAL-$5^i$ and COCO-$20^i$ datasets, and achieved high performance in the few-shot segmentation task, outperforming the previous state-of-the-art model by a large margin ($6.3\%$ and $9.5\%$ in mIoU, respectively). Furthermore, VLP-SAM demonstrates its generality in unseen objects that are not included in the training data. Our code is available at \href{https://github.com/kosukesakurai1/VLP-SAM}{https://github.com/kosukesakurai1/VLP-SAM}.
\end{abstract}

\vspace{3mm}
\vspace{8mm}
]


\section{Introduction}

\begin{figure*}[ht]
\centering
\includegraphics[width=0.9\linewidth]{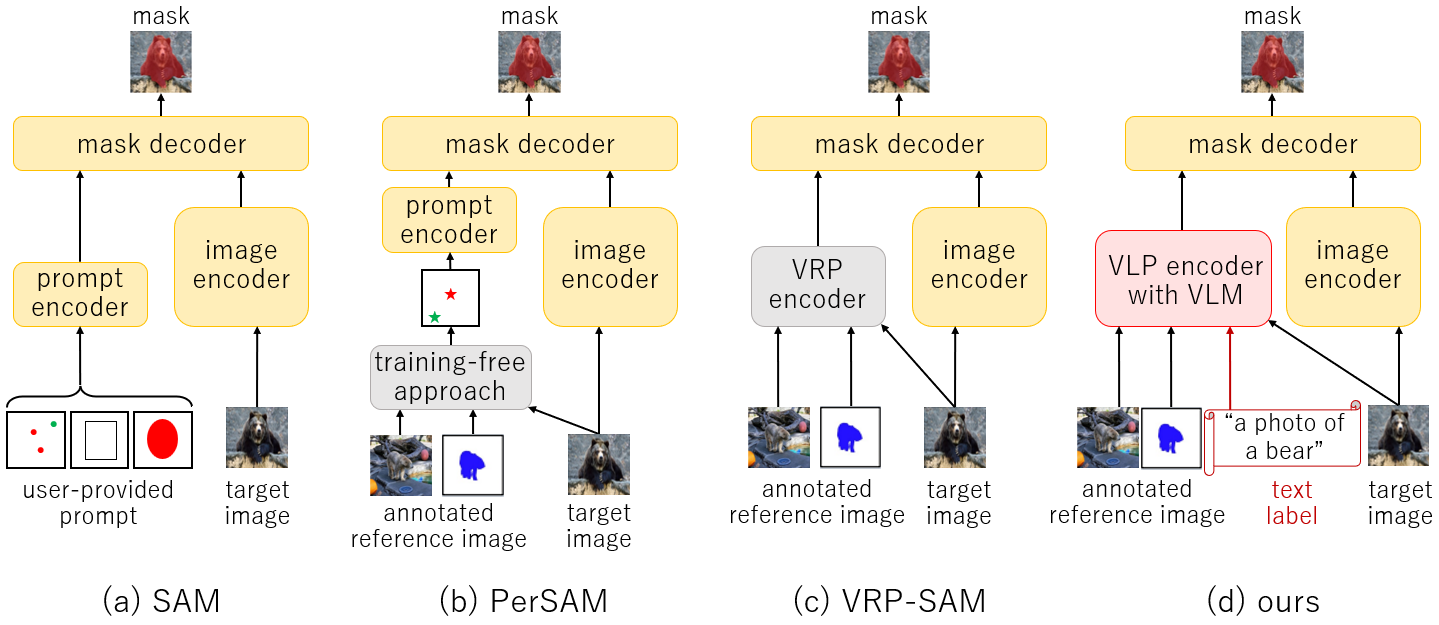}
\caption{Comparison of our proposed method VLP-SAM and three previous models. (a) SAM~\cite{SAM} is a category-agnostic interactive zero-shot segmentation model that inputs user-provided prompts for each target image. (b) PerSAM~\cite{PerSAM} is a training-free SAM-based few-shot segmentation model that inputs positive and negative points on the target image into SAM, without requiring user-provided prompts. (c) VRP-SAM~\cite{VRP-SAM} is a SAM-based few-shot segmentation model incorporating a VRP encoder to generate prompt embeddings for SAM from an annotated reference image. (d) Our proposed method, VLP-SAM, introduces a novel prompt encoder for SAM, called the VLP encoder, which accepts both reference images and text labels as input via a vision-language model.}
\label{fig:2}
\end{figure*}

In recent years, Segment Anything Model (SAM)~\cite{SAM} has been released as a foundational image segmentation model trained on a large dataset with billion mask labels. SAM has powerful zero-shot capabilities that can segment any object in a target image by taking user-provided prompts consisting of points, bounding boxes, or coarse masks. However, user-provided prompts require the user's comprehensive understanding of the target objects, and different customized prompts are needed for each target image. Furthermore, SAM outputs category-agnostic masks, limiting its usability in real-world applications.

Few-shot segmentation models~\cite{PerSAM, Mather, VRP-SAM, APSeg} addressed these issues by inputting a few annotated reference images as prompts to SAM and can segment specific objects in target images without user-provided prompts. Based on the pixel-level similarity between annotated reference images and target images, few-shot segmentation models segment target objects of the same category as the reference images. On the other hand, previous SAM-based few-shot segmentation models only use annotated reference images as prompts, resulting in limited accuracy due to the lack of reference information.

In this work, we propose a novel few-shot segmentation model, \textit{Vision and Language reference Prompt into SAM} (VLP-SAM), that inputs not only reference images but also text labels as prompts to SAM. By using text information, VLP-SAM can leverage not only the visual similarity between annotated reference images and target images but also the semantic similarity between text labels and target objects. 

In particular, VLP-SAM is a simple and scalable structure that trains a new SAM's prompt encoder with minimal learnable parameters that outputs prompt embedding from a target image, an annotated reference image, and a text label. First, VLP-SAM embeds images (a target image and a reference image) and language (a text label) in the same embedding space using a multimodal vision-language model (VLM). Instead of CLIP~\cite{CLIP}, which is trained only on CLS tokens, a model that has pixel-level similarity between images and texts (e.g., CLIP Surgery~\cite{CLIP-Surgery}) is used as the VLM. Then, a visual prototype that aggregates the features of the target object from the annotated reference image and a textual prototype that aggregates semantic information of the target object from the text embedding are concatenated to each embedding. Furthermore, following previous work~\cite{VRP-SAM}, a pseudo mask of the target image created from the annotated reference image and an attention mask of the target image created from the text label are added to improve accuracy. Finally, VLP-SAM outputs prompt embeddings for SAM from learnable queries via Transformer-based attention~\cite{Transformer} interacting with each embedding. The prompt embeddings created by the VLP-SAM that contain visual and semantic information about the target object are input to the SAM's mask decoder, allowing the VLP-SAM to segment target objects without user-provided prompts.

Our proposed method, VLP-SAM, is a highly accurate and general few-shot segmentation model that can segment any class not included in the training data by leveraging the benefits of large-scale models of SAM and VLM. Furthermore, VLP-SAM is a scalable SAM-based segmentation model that leverages a new prompt encoder that can input a variety of modalities, such as reference images and text, rather than traditional SAM prompts, such as points and bounding boxes. 

To demonstrate the effectiveness of VLP-SAM, we conducted experiments on the PASCAL-$5^i$~\cite{PASCAL} and COCO-$20^i$~\cite{COCO} datasets. Our experimental results show that VLP-SAM outperforms the previous state-of-the-art model~\cite{VRP-SAM} by a large margin ($6.3\%$ and $9.5\%$ in mIoU, respectively) in a few-shot segmentation task. Furthermore, VLP-SAM demonstrates its generality in unseen objects.

The main contributions are summarized as follows:
\begin{itemize}
    \item We propose a novel few-shot segmentation model, \textit{Vision and Language reference Prompt into SAM} (VLP-SAM), which inputs not only reference images but also text labels as prompts to SAM. Our method is a highly accurate model that can utilize the visual information of reference images and the semantic information of text labels.
    \item VLP-SAM introduces a novel SAM's prompt encoder using VLM with pixel-text matching, achieving a scalable SAM-based model with minimal learnable parameters.
    \item We demonstrate that VLP-SAM outperforms the previous state-of-the-art model by a large margin in unseen objects.
\end{itemize}

\section{Related Work}
\subsection{Segment Anything Model}
Segment Anything Model (SAM)~\cite{SAM} is a category-agnostic interactive segmentation model trained on a large-scale SA-1B dataset containing over 1 billion masks. SAM shows powerful zero-shot capabilities to new objects without additional training by taking user-provided prompts consisting of points, bounding boxes, or coarse masks. Due to its generality, SAM has been applied to a variety of research~\cite{HQSAM, MedSAM, FastSAM, Semantic-sam, Grounded-sam, Depth-anything, Samrs, survey-sam}. As depicted in \cref{fig:2}(a), SAM comprises three modules: an image encoder, a prompt encoder, and a mask decoder. The image encoder is a Vision Transformer~\cite{ViT} backbone to extract image embeddings. The prompt encoder generates prompt embeddings from geometric prompts such as points and boxes. The mask decoder is a Transformer-based decoder~\cite{Transformer} that outputs a class-agnostic mask from image embeddings and prompt embeddings. While SAM can segment any object in zero-shot, it requires user-provided prompts for each target image, meaning human interaction and knowledge of the target image are needed. Furthermore, SAM does not output the class labels for each mask, limiting its usability in real-world applications.

\subsection{Few-shot Segmentation Model with SAM}
Few-shot segmentation models~\cite{Panet, PGMA-Net, Seggpt, BAM, RefLDM, Segic, PI-CLIP, SINE, survey-few} aim to segment objects in target images belonging to the same category as annotated reference images. Specifically, SAM-based few-shot segmentation models~\cite{PerSAM, VRP-SAM, Mather, APSeg, SAMIC, Alignsam, PDM, NubbleDrop, PMC, Foreground-Covering} only take a few reference image-mask pairs as prompts, instead of user-provided prompts for each target image. This allows leveraging the richness of large-scale foundational model SAM while addressing SAM's weaknesses: user interaction and class-agnostic masks.

SAM-based few-shot segmentation models can be mainly classified into two types: training-free models~\cite{PerSAM, Mather, PDM, NubbleDrop, PMC} that input geometric prompts derived from annotated reference images into SAM's prompt encoder, and meta-learning models~\cite{VRP-SAM, APSeg, Foreground-Covering} that introduce the new SAM's prompt encoder and input prompt embeddings into SAM's mask decoder. Training-free models generate geometric prompts of target objects from pixel-level correlations between annotated reference images and target images without user interaction. For example, as shown in \cref{fig:2}(b), PerSAM~\cite{PerSAM} selects the most positive and negative points from pixel-level correlations to input into SAM as prompts, and segments target objects without training. However, the performance of training-free models heavily depends on the quality of pseudo masks generated from pixel-level correlations, leading to incorrect prompts and reduced accuracy. Additionally, training-free models limit their scalability because SAM only accepts points, boxes, and masks as prompts.

Meta-learning models generate prompt embeddings for SAM derived from annotated reference images instead of geometric prompts via a meta-learning procedure. As depicted in \cref{fig:2}(c), VRP-SAM~\cite{VRP-SAM} introduces a novel prompt encoder, the VRP encoder, which outputs prompt embeddings from annotated reference and target images. These prompt embeddings are then input into SAM's mask decoder, creating a scalable few-shot segmentation model capable of predicting unseen classes not included in the training data. In this study, as shown in \cref{fig:2}(d), we introduce a novel SAM's prompt encoder (VLP encoder) that leverages a vision-language model (VLM) to capture both visual and semantic information for higher segmentation accuracy.

\subsection{Multimodal Vision-Language Model}
Multimodal vision-language models (VLM) embed images and text in a unified embedding space, and various VLMs have been released~\cite{CLIP, CoCa, ALIGN, BLIP, G-DINO, FIS, PFIS}. A representative model, CLIP~\cite{CLIP}, is a foundational model learned through contrastive learning~\cite{contrastive, Vilbert} on 400 million image-text pairs. Its zero-shot capability has been widely utilized in a variety of computer vision tasks~\cite{StableDiffusion, CoOp, GLIP, Clip4clip, Wav2clip, LARE, PointCLIP}. However, for object detection and image segmentation tasks, pixel-text matching rather than CLS token-text matching like CLIP is required to capture object-level features. VLM with pixel-text matching~\cite{CLIP-Surgery, OWL-ViT, Denseclip, Zegclip} can capture spatial and semantic information of objects from each patch-text relationship. Notably, CLIP Surgery~\cite{CLIP-Surgery} enhances the class attention map (CAM)~\cite{cam, Grad-cam, Clims, CLIP-ES} of CLIP to create high-performance pixel-text correlations (attention maps) without additional training. This model retains the benefits of CLIP's large-scale foundational model while providing image-text embeddings that capture object-level semantic features. Therefore, in this study, we employ CLIP Surgery as VLM and CLIP as a comparison.

\section{Methodology: VLP-SAM}

\begin{figure*}[ht]
\centering
\includegraphics[width=0.85\linewidth]{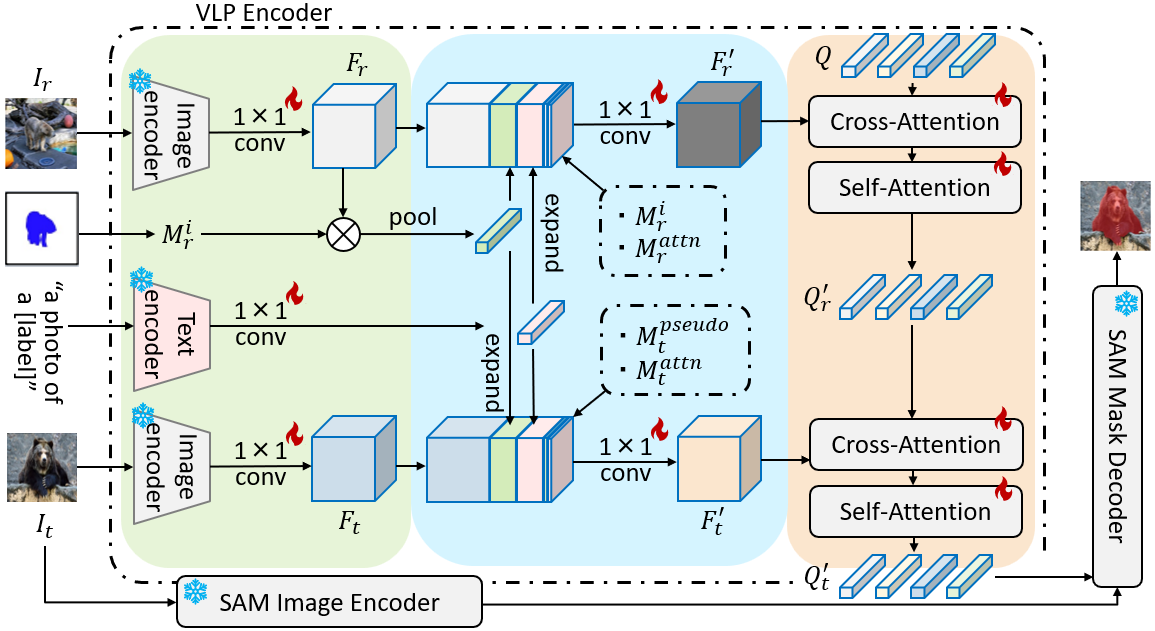}
\caption{Overview of our proposed method VLP-SAM. VLP-SAM introduces a novel prompt encoder for SAM, called the VLP encoder, which generates the prompt embedding $\bm{Q}'_t$ with vision-language information. In particular, the VLP encoder first embeds the target image $I_t$, the reference image $I_r$, and the text label into the unified embedding space using the VLM encoder. These embeddings are then concatenated with prototypes and masks that aggregate information about the target object and the text label. Finally, the prompt embedding $\bm{Q}'_t$, refined through Transformer-based attention, is fed into SAM's mask decoder, enabling VLP-SAM to produce a more accurate mask of the target object.}
\label{fig:3}
\end{figure*}

In this paper, we propose a novel few-shot segmentation model, \textit{Vision and Language reference Prompt into SAM} (VLP-SAM), that inputs not only reference images but also text labels as prompts to SAM. By using text labels with VLM, VLP-SAM can utilize both the visual information of reference images and the semantic information of text labels. 

An overview of VLP-SAM is shown in \cref{fig:3}. A novel SAM's prompt encoder, VLP encoder, generates prompt embeddings from the target image, the annotated reference image, and the text label with minimal learnable parameters. By inputting these prompt embeddings with vision-language information into SAM's mask decoder, VLP-SAM outputs the mask of the target object. The details of the VLP encoder and the training process are described below.

\subsection{VLP Encoder}
In the few-shot segmentation task, the target image $I_t$, the reference image $I_r$, and its annotation mask $M^i_r$, where $i$ denotes the category of the annotated object, are given. Additionally, this study prepares the class label of the annotated object $i$ and uses it as a text $T_i$, such as ``a photo of a [$i$].'' By utilizing the information from the annotated reference image $I_r + M^i_r$ and the text label $T_i$, VLP-SAM can accurately predict the mask $M^i_t$ of the category $i$ in the target image. 

First, using a multimodal VLM, $I_r$, $I_t$, and $T_i$ are embedded into the same latent space, and image embeddings $\bm{F}_r \in \mathbb{R}^{C\times{H}\times{W}}$, $\bm{F}_t \in \mathbb{R}^{C\times{H}\times{W}}$, text embedding $\bm{F}_\text{text} \in \mathbb{R}^{C\times{1}}$ are obtained via a $1\times{1}$ convolution layer~\cite{CNN}. Since image embeddings use not only CLS tokens but also $H\times{W}$ patches, CLIP Surgery~\cite{CLIP-Surgery}, which considers pixel-text matching, is used instead of CLIP~\cite{CLIP}. To prevent overfitting of the VLP encoder, the VLM is frozen during the training phase. Next, the visual prototype $\bm{P}_i$ that aggregates the features of the target object $i$ from $\bm{F}_r$ and $M^i_r$, and the textual prototype $\bm{F}_\text{text}$ that aggregates the semantic information of category $i$ are extracted. The visual prototype $\bm{P}_i$ is the average embedding of the target object from the reference image $\bm{F}_r$, and it is formulated as follows:

\begin{equation}
    \bm{P}_i={\rm MaskAvgPool}(\bm{F}_r, M^i_r).
    \label{eq:1}
\end{equation}

Furthermore, following previous work~\cite{VRP-SAM}, the masks of the reference and target images are utilized to enhance the information about objects of category $i$. The mask of the reference image uses the annotation mask $M^i_r$, while the mask of the target image uses a pseudo mask $M^{pseudo}_t$, which is obtained through a common training-free approach. The pseudo mask $M^{pseudo}_t \in \mathbb{R}^{H\times{W}}$ is created by the cosine similarity correlations $\bm{S}\in\mathbb{R}^{(H\times{W})\times{(H\times{W})}}$ between the target image and the annotated reference image. Additionally, following CLIP Surgery~\cite{CLIP-Surgery}, the attention masks $M^{attn}_r, M^{attn}_t \in \mathbb{R}^{H\times{W}}$ of the reference and target images are utilized. The attention masks are derived from the cosine similarity between the image and text embeddings, and they include semantic information with the text label. Enhanced image embeddings $\bm{F}'_r\in\mathbb{R}^{C\times{H}\times{W}}$, $\bm{F}'_t\in\mathbb{R}^{C\times{H}\times{W}}$ are obtained by concatenating the visual prototype $\bm{P}_i$, textual prototype $\bm{F}_\text{text}$, pseudo mask $M^i_r$, $M^{pseudo}_t$, and attention mask $M^{attn}_r$, $M^{attn}_t$ with $\bm{F}_r$ and $\bm{F}_t$ through a $1\times{1}$ convolution layer to reduce the dimension. These enhanced embeddings $\bm{F}'_r$, $\bm{F}'_t$ are formulated as follows:

\begin{align}
    \bm{F}'_r &= {\rm Conv}({\rm concat}(\bm{F}_r, \bm{P}_i, \bm{F}_\text{text}, M^i_r, M^{attn}_r)), \label{eq:2} \\
    \bm{F}'_t &= {\rm Conv}({\rm concat}(\bm{F}_t, \bm{P}_i, \bm{F}_\text{text}, M^{pseudo}_t, M^{attn}_t)). \label{eq:3}
\end{align}

\begin{table*}[ht]
\begin{center}
\caption{Performance of one-shot segmentation on PASCAL-$5^i$ and COCO-$20^i$. Backbone indicates the image encoder within the novel SAM's prompt encoder (VRP and VLP encoders). VLP-SAM outperforms the previous state-of-the-art model VRP-SAM (ResNet-50) by a large margin ($6.3\%$ and $9.5\%$ in mIoU, respectively).}
\label{tab:1}
\scalebox{0.95}{
\begin{tabular}{@{}ccccccccccccc@{}}
\toprule
\multirow{2}{*}{\makecell[c]{\textbf{Method}}} & \multirow{2}{*}{\makecell[c]{\textbf{Backbone}}} & \multirow{2}{*}{\makecell[c]{\textbf{\# Learnable}\\\textbf{params}}} & \multicolumn{5}{c}{\textbf{PASCAL-$5^i$}} & \multicolumn{5}{c}{\textbf{COCO-$20^i$}} \\ \cmidrule(lr){4-8} \cmidrule(lr){9-13}
& & & F-0 & F-1 & F-2 & F-3 & \textbf{\underline{Mean}} & F-0 & F-1 & F-2 & F-3 & \textbf{\underline{Mean}} \\ \midrule
\multirow{4}{*}{VRP-SAM} & ResNet-50 & 1.6M & 73.18 & 75.74 & 69.50 & 64.21 & 70.66 & 43.40 & 54.43 & 53.28 & 50.53 & 50.41\\
& CLIP (ResNet-50) & 1.5M & 64.44 & 70.38 & 64.44 & 57.72 & 64.25 & 30.45 & 41.51 & 43.74 & 40.06 & 38.94\\
& CLIP (ViT-B/16) & 1.3M & 68.64 & 72.33 & 65.78 & 60.38 & 66.78 & 37.56 & 48.11 & 50.84 & 46.83 & 45.84\\
& CLIP Surgery (ViT-B/16) & 1.3M & 70.50 & 75.88 & 65.44 & 61.63 & 68.36 & 40.62 & 52.28 & 52.77 & 49.87 & 48.89\\ \midrule
\multirow{2}{*}{\makecell[c]{VLP-SAM\\(ours)}} & CLIP (ViT-B/16) & 1.4M & 71.92 & 73.12 & 65.28 & 62.20 & 68.13 & 41.74 & 52.76 & 52.49 & 51.66 & 49.66\\
& CLIP Surgery (ViT-B/16) & 1.4M & \textbf{76.94} & \textbf{83.08} & \textbf{72.73} & \textbf{75.27} & \textbf{77.01} & \textbf{52.40} & \textbf{63.90} & \textbf{62.10} & \textbf{61.26} & \textbf{59.92}\\
\bottomrule
\end{tabular}
}
\end{center}
\end{table*}

Finally, the VLP encoder generates prompt embeddings for SAM using enhanced image embeddings $\bm{F}'_r$, $\bm{F}'_t$ and learnable queries $\bm{Q}\in\mathbb{R}^{N\times{C}}$, where $N$ denotes the number of tokens in the prompt embeddings. The learnable queries $\bm{Q}$ initially interact with the reference features $\bm{F}'_r$ through cross-attention and self-attention layers to acquire category-specific information. Subsequently, these queries interact with the target features $\bm{F}'_t$ to acquire foreground information in the target image. These processes are formulated as follows:

\begin{align}
    \bm{Q}'_r &= {\rm SelfAttn}({\rm CrossAttn}(\bm{Q}, \bm{F}'_r)), \label{eq:4} \\
    \bm{Q}'_t &= {\rm SelfAttn}({\rm CrossAttn}(\bm{Q}'_r, \bm{F}'_t)). \label{eq:5}
\end{align}

The final $\bm{Q}'_t$ represents the prompt embeddings for SAM, containing both visual and semantic information of the target object.

\subsection{Training}
The prompt embeddings obtained from the VLP encoder are input to SAM's mask decoder instead of geometric prompts. VLP-SAM can predict the mask $M^i_t$ for the category $i$ of the target image from the mask decoder by utilizing the reference images and the text labels. We employ Binary Cross-Entropy (BCE) loss and Dice loss to train the VLP encoder. The loss of VLP-SAM is formulated as follows:

\begin{equation}
\begin{aligned}
{\rm Loss} & =\underbrace{-\frac{1}{n} \sum_{j=1}^n\left[y_j \log \left(p_j\right)+\left(1-y_j\right) \log \left(1-p_j\right)\right]}_{\text {BCE Loss }} \\
& +\underbrace{1-\frac{2 \sum_{j=1}^n\left(p_j \cdot y_j\right) +1}{\sum_{j=1}^n\left(p_j+y_j\right)+1}}_{\text {Dice Loss }}.
\end{aligned}
\end{equation}
where $n$ represents the total number of pixels, $y_j$ denotes the pixel $j$ of the ground truth mask $M^i_{gt}$, and $p_j$ denotes the pixel $j$ of the predicted mask $M^i_t$. During training, the parameters of SAM's image encoder, mask encoder, and VLM encoder are frozen to prevent overfitting. By leveraging the weights of pre-trained foundational models, VLP-SAM enables few-shot segmentation into any classes not included in the training data.

\section{Experiment}
\subsection{Settings}
To demonstrate the effectiveness of VLP-SAM, we conducted experiments on the PASCAL-$5^i$~\cite{PASCAL} and COCO-$20^i$~\cite{COCO} datasets following the few-shot setting~\cite{VRP-SAM, BAM, CyCTR}. We validate the generalization capability of VLP-SAM by dividing both datasets into four folds: three folds are used for training, and the remaining one fold is used for testing. PASCAL-$5^i$ consists of 20 classes (15 for training and 5 for testing), while COCO-$20^i$ consists of 80 classes (60 for training and 20 for testing), with no class overlap between training and testing. In each fold, 1,000 reference-target pairs are randomly sampled for testing. 

We compare VLP-SAM with a SAM-based few-shot segmentation model, VRP-SAM~\cite{VRP-SAM}, which does not input text labels. We use ResNet-50~\cite{ResNet}, CLIP~\cite{CLIP} (ViT-B/16~\cite{ViT}, ResNet-50), CLIP Surgery~\cite{CLIP-Surgery} (ViT-B/16) as the image encoder backbone. ResNet-50 is a pre-trained model on ImageNet~\cite{ImageNet}, and is only used in VRP-SAM because it cannot accept text as input. To ensure a fair comparison, all methods employ the AdamW optimizer~\cite{AdamW} with an initial learning rate of 1e-4, a batch size of 8, 100 epochs for PASCAL-$5^i$, 50 epochs for COCO-$20^i$, and the number of learnable queries of 50. In addition, VLP-SAM sets the input text to the VLM as ``a photo of a [label].'' Following previous work~\cite{VRP-SAM, Mather, HDMNet}, the evaluation metric is mean intersection over union (mIoU).

\subsection{Results}
\cref{tab:1} shows the performance of one-shot segmentation on PASCAL-$5^i$ and COCO-$20^i$. As shown in \cref{tab:1}, VLP-SAM (CLIP Surgery) achieves mIoU scores of 77.01 on PASCAL-$5^i$ and 59.92 on COCO-$20^i$ with 1.4M learnable parameters. Furthermore, VLP-SAM (CLIP Suegery) outperforms the previous state-of-the-art model VRP-SAM (ResNet-50) by a large margin ($6.3\%$ and $9.5\%$, respectively). Additionally, compared to the text-free model VRP-SAM (CLIP Surgery) with the same backbone, our method achieves superior performance by increasing 8.6 and 11.0 mIoU on the respective datasets. These results demonstrate the high accuracy of VLP-SAM for few-shot segmentation and its generalization capability to unseen classes.

\section{Discussion}
This section discusses the effectiveness of the proposed method VLP-SAM. The proposed method significantly outperforms the previous state-of-the-art model by inputting not only reference images but also text labels as prompts to SAM. This improvement is attributed to its semantic similarities with text labels that lead to segmenting complex objects such as ``dog-like cats'' or ``fully clothed people.'' Furthermore, by leveraging prompt embeddings with multiple perspectives, such as visual and semantic features of target objects, VLP-SAM demonstrates powerful generalization capability to new objects.

In addition, VLP-SAM is a scalable SAM-based few-shot segmentation model with minimal learnable parameters. Traditional SAM only accepts geometric prompts such as points, boxes, or masks. On the other hand, by introducing a novel SAM's prompt encoder with VLM, VLP-SAM removes the limitations of geometric prompts, enabling the input of any modality like text. The introduction of text will expand to applications such as prompt engineering and open vocabulary.



\section{Conclusion and Future Work}
In this paper, we proposed VLP-SAM, a novel SAM-based few-shot segmentation model that inputs not only reference images but also text labels as prompts. By learning a novel prompt encoder for SAM using a multimodal VLM, VLP-SAM achieves high accuracy and generality. In particular, experiments on two datasets demonstrated that VLP-SAM outperforms previous state-of-the-art models by a large margin. Additionally, VLP-SAM proved its generalization to novel objects, making it applicable to various scenarios.

In future work, we will focus on enhancing and reducing reference information. By leveraging the simple and scalable structure of the VLP encoder, it is possible to add various modals of reference information, such as 3D images and depth information as input to SAM. Conversely, by exploiting the richness of large-scale foundational models like SAM and VLM, reducing annotation masks or reference images could enhance usability by enabling segmentation with minimal input information.

\section*{Acknowledgment}
This work was supported by JSPS, Japan KAKENHI Grant Numbers 21H04600 and 24H00370.

\section*{Declaration of Competing Interest}
The authors declare that they have no known competing financial interests or personal relationships that could have appeared to influence the work reported in this paper.

\bibliographystyle{unsrt}  
\bibliography{references}  



\end{document}